\documentstyle[11pt,a4,colacl,macros,other-abstract]{article}

 \include{epsf}

 \title{On the Evaluation and Comparison of Taggers: the Effect of
  Noise in Testing Corpora.}
 \author{{\bf Llu\'{\i}s Padr\'o} \and {\bf Llu\'{\i}s M\`arquez}\\
 Dep. LSI. Technical University of Catalonia\\
 c/ Jordi Girona 1-3. 
 08034 Barcelona\\
 {\tt \{padro,lluism\}@lsi.upc.es}}

\begin{document}

\maketitle
\bibliographystyle{acl}

\begin{abstract}
This paper addresses the issue of {\sc pos} tagger evaluation.
Such evaluation is usually performed by comparing the tagger
output with a reference test corpus, which is assumed to
be error-free. Currently used
corpora contain noise which causes the obtained performance
to be a distortion of the real value. We analyze 
to what extent this distortion may invalidate the comparison
between taggers or the measure of the improvement given by a new system. 
The main conclusion is that a more rigorous testing experimentation 
setting/designing is needed to reliably evaluate and compare tagger
accuracies.
\end{abstract}

\section{Introduction and Motivation}
\label{seccio-introduccio}

Part of Speech {\sc (pos)} Tagging is a quite well defined 
{\sc nlp} problem, which consists of
assigning to each word in a text the proper morphosyntactic tag
for the given context. Although many words are
ambiguous regarding their {\sc pos}, in most cases they can be 
completely disambiguated taking into account an adequate context.
Successful taggers have been built using several approaches,
such as statistical techniques, symbolic machine learning 
techniques, neural networks, etc.
The accuracy reported by most current taggers ranges from 96--97\%
to almost 100\% in the linguistically--motivated 
Constraint Grammar environment. 

Unfortunately, there have been very few direct comparisons of
alternative taggers\footnote{One of the exceptions is the work
  by~\cite{Samuelsson97}, in which a very strict comparison between
  taggers is performed.} on identical test data. 
However, in most current papers it is argued that the performance of
some taggers is better than others as a result of some kind of
indirect comparisons between them.
We think that there are a number of not enough
controlled/considered factors that make these conclusions 
dubious in most cases.

In this direction, the present paper aims to point out 
some of the difficulties arising when evaluating and comparing
tagger performances against a reference test corpus, and to make 
some criticism about common practices followed by
the {\sc nlp} researchers in this issue.

  The above mentioned factors can affect either the evaluation 
or the comparison process. Factors affecting the evaluation process
are:
(1) Training and test experiments are usually performed over 
noisy corpora which distorts the obtained results,
(2) performance figures are too often calculated from only a
single or very small number of trials, though average results from multiple
trials are crucial to obtain reliable estimations of 
accuracy \cite{Mooney96}, (3) testing experiments 
are usually done on corpora with the same characteristics as the
training data --usually a small fresh portion of the training
corpus-- but no serious attempts have been done in order 
to determine the reliability of the results when moving from 
one domain to another \cite{Krovetz97}, and (4)
no figures about computational effort --space/time complexity-- 
are usually reported, even from an empirical perspective.
 A factors affecting the comparison process is that
comparisons between taggers are often indirect, while
they should be compared under the same conditions
in a multiple--trial experiment with statistical tests of
significance.

For these reasons, this paper calls for
a discussion on {\sc pos} taggers evaluation, aiming
to establish a more rigorous test experimentation 
setting/designing, indispensable to extract 
reliable conclusions. As a starting point, we will focus 
only on how the noise in the test corpus can affect the 
obtained results.

\section{Noise in the testing corpus}
\label{seccio-soroll}

From a machine learning perspective, the relevant noise in the corpus
is that of non systematically mistagged words
(i.e. different annotations for words appearing in the same
syntactic/semantic contexts). 

Commonly used annotated corpora have noise.
See, for instance, the following examples from the Wall
Street Journal {\sc (wsj)} corpus:

Verb participle forms are sometimes tagged as such 
{\sc (vbn)} and also as adjectives {\sc (jj)}
in other sentences with no structural differences:
{\small
\begin{itemize}
\item[1a)]{\tt... failing\_VBG to\_TO voluntarily\_RB submit\_VB 
the\_DT {\em requested\_VBN} information\_NN ...}

\item[1b)]{\tt... a\_DT large\_JJ sample\_NN of\_IN {\em married\_JJ} women\_NNS 
with\_IN at\_IN least\_JJS one\_CD child\_NN ...}
\end{itemize}}
 Another structure not coherently tagged are noun
chains when the nouns {\sc (nn)} are ambiguous and can be also
adjectives {\sc (jj)}:
{\small
\begin{itemize}
\item[2a)]{\tt ... Mr.\_NNP Hahn\_NNP ,\_, the\_DT 62-year-old\_JJ chairman\_NN 
and\_CC {\em chief\_NN executive\_JJ officer\_NN} of\_IN 
Georgia-Pacific\_NNP Corp.\_NNP ...}

\item[2b)]{\tt... Burger\_NNP King\_NNP 's\_POS {\em chief\_JJ executive\_NN officer\_NN} ,\_, 
Barry\_NNP Gibbons\_NNP ,\_, stars\_VBZ in\_IN ads\_NNS saying\_VBG ...}
\end{itemize}}

The noise in the test set produces a wrong estimation of 
accuracy, since correct answers are computed as wrong and 
vice-versa. 
In following sections we will show how this uncertainty 
in the evaluation may be, in some cases, larger than 
the reported improvements from one system to another, 
so invalidating the conclusions of the comparison.

\section{Model Setting}
\label{seccio-plantejament}

  To study the appropriateness of the choices made by a {\sc pos} tagger, a 
reference tagging must be selected and assumed to be correct in order to
compare it with the tagger output. This is usually done by assuming that 
the disambiguated test corpora being used contains the right {\sc pos} disambiguation.
This approach is quite right when the tagger error rate is larger enough than
the test corpus error rate, nevertheless, the current {\sc pos} taggers have reached 
a performance level that invalidates this choice, since the tagger error rate is 
getting too close to the error rate of the test corpus.

  Since we want to study the relationship between the tagger error rate and
the test corpus error rate, we have to establish an absolute reference point.
Although \cite{Church92} questions the concept of {\sl correct analysis}, 
\cite{Samuelsson97} 
establish that there exists a --statistically significant-- 
{\sl absolute} correct disambiguation, respect 
to which the error rates of either the tagger or the test corpus can be computed. 
What we will focus on is how distorted is the tagger error rate by the use of a 
noisy test corpus as a reference.  

  The cases we can find when evaluating the performance of a certain tagger are 
presented in table \ref{taula-cassos}. {\sc ok}/$\neg${\sc ok} stand
for a right/wrong tag (respect to the absolute correct disambiguation). When
both the tagger and the test corpus have the correct tag, the tag is correctly
evaluated as {\sl right}. When the test corpus has the correct tag and the tagger gets it
wrong, the occurrence is correctly evaluated as {\sl wrong}. But problems arise when
the test corpus has a wrong tag: If the tagger gets it correctly, it is 
evaluated as {\sl wrong} when it should be {\sl right} (false negative). If the 
tagger gets it wrong, it will be rightly evaluated as {\sl wrong} if the error 
commited by the tagger is other than the error in the test corpus, but wrongly 
evaluated as {\sl right} (false positive) if the error is the same. 
\begin{table}[htb] \centering
\begin{tabular}{|cc|c|c|} \hline
  {\small corpus}    & {\small tagger}    & {\small eval: {\sl right}} & {\small eval: {\sl wrong}}\\\hline
  {\sc ok}$_c$       & {\sc ok}$_t$       & {\small $(1\menys C)t$}    &   --\\ 
  {\sc ok}$_c$       & $\neg${\sc ok}$_t$ & --                & {\small $(1\menys C)(1\menys t)$}\\\hline
  $\neg${\sc ok}$_c$ & {\sc ok}$_t$       & --                & {\small $C u$}\\ 
  $\neg${\sc ok}$_c$ & $\neg${\sc ok}$_t$ & {\small $C(1\menys u)p$}   & {\small $C(1\menys u)(1\menys p)$}\\\hline
\end{tabular}
\caption{\small{Possible cases when evaluating a tagger.}}
\label{taula-cassos}
\end{table}
  Table~\ref{taula-cassos} shows the computation of the percentages of each case. 
The meanings of the used variables are:
\begin{itemize}
{\parskip 0pt
\item[$C$:]Test corpus error rate. Usually an estimation is supplied with
           the corpus.
\item[$t$:]Tagger performance rate on words rightly tagged in the test corpus.
           It can be seen as $P(${\sc ok}$_t|${\sc ok}$_c)$.
\item[$u$:]Tagger performance rate on words wrongly tagged in the test corpus.
           It can be seen as $P(${\sc ok}$_t|\neg${\sc ok}$_c)$.
\item[$p$:]Probability that the tagger makes the same error as the test
          corpus, given that both get a wrong tag.
\item[$x$:]{\sl Real} performance of the tagger, i.e. what would be obtained 
           on an error--free test set.
\item[$K$:]Observed performance of the tagger, computed on the noisy
  test corpus.
}
\end{itemize}

For simplicity, we will consider only performance on ambiguous 
words. Considering unambiguous words will make the analysis more 
complex, since it should be taken into account that neither the behaviour of 
the tagger (given by $u$, $t$, $p$) nor the errors in the test corpus (given by $c$)
are the same on ambiguous and unambiguous words. Nevertheless, this is an issue that
must be further addressed. 
 
 If we knew each one of the above proportions, we would be able to compute the {\sl real}
performance of our tagger ($x$) by adding up the {\sc ok$_t$} rows from
table~\ref{taula-cassos}, i.e. the cases in which the tagger 
got the right disambiguation independently from the tagging of the test set:
\vspace{-0.25cm}
\begin{equation}
\label{eq-x}
 x \igual (1\menys C)t \mes  C u 
\end{equation}

\vspace{-0.2cm}
 The equation of the observed performance can also
be extracted from table \ref{taula-cassos}, adding up what is evaluated
as {\sl right}:
\vspace{-0.25cm}
\begin{equation}
\label{eq-k}
 K \igual (1\menys C)t \mes  C(1\menys u)p 
\end{equation}

\vspace{-0.2cm}
 The relationship between the real and the observed performance
is derived from \ref{eq-x} and \ref{eq-k}:
\vspace{-0.25cm}
$$
 x \igual K \menys  C(1\menys u)p \mes  C u 
\vspace{-0.2cm}
$$
 Since only $K$ and $C$ are known (or approximately estimated) we can
not compute the real performance of the tagger. All we can do is to 
establish some reasonable bounds for $t$, $u$ and $p$, and see in which
range is $x$. 

  Since all variables are probabilities, they are bounded in $[0,1]$. We also can 
assume\footnote{In the cases we are interested in --that is, current
                 systems-- the tagger 
                 observed performance, $K$, is over 90\%, while 
                 the corpus error rate, $C$, is below 10\%.}
that $K>C$. We can use this constraints and the above equations to bound 
the values of all variables. From~\ref{eq-k}, we obtain:

\vspace{-0.3cm}
$$
\hspace{-0.2cm}
{\scriptsize
\begin{array}{lll}
{\displaystyle u = 1 \menys \frac{K\menys t(1\menys C)}{C p}};&
{\displaystyle p = \frac{K\menys t(1\menys C)}{C(1\menys u)}};&
{\displaystyle t = \frac{K\menys C(1\menys u)p}{1\menys C}}
\end{array}}
$$

  Thus, $u$ will be maximum when $p$ and $t$ are maximum (i.e. $1$). 
This gives an upper bound for $u$ of $(1\menys K)/C$. When $t\igual
0$, $u$ will range in $[-\infty,1\menys K/C]$ depending on the value 
of $p$. Since we are assuming $K\mg C$, the most informative lower 
bound for $u$ keeps being zero. Similarly,  $p$ is minimum when 
$t\igual 1$ and $u\igual 0$. When $t\igual 0$ the value for $p$ will 
range in $[K/C,+\infty]$ depending on $u$. Since $K\mg C$, the most 
informative upper bound for $p$ is still $1$. Finally, $t$ will be 
maximum when $u \igual  1$ and $p \igual  0$, and minimum when
$u\igual 0$ and $p\igual 1$. Summarizing:
{\footnotesize
\begin{eqnarray}
\label{eq-ubounds}
 \hspace*{-7mm}{\displaystyle 0} & {\displaystyle\!\!\!\!\mpi u\mpi\!\!\!\!} & {\displaystyle min\left\{1,\frac{1\menys K}{C}\right\}}\vspace*{2mm}\\
\label{eq-pbounds}
 \hspace*{-7mm}{\displaystyle max\left\{0,\frac{K\mes C\menys 1}{C}\right\}} & {\displaystyle\!\!\!\!\mpi p\mpi\!\!\!\!} & {\displaystyle 1}\vspace*{2mm}\\
\label{eq-tbounds}
 \hspace*{-7mm}{\displaystyle\frac{K\menys C}{1\menys C}} & {\displaystyle\!\!\!\!\mpi t\mpi\!\!\!\!} & {\displaystyle min\left\{1,\frac{K}{1\menys C}\right\}}
\end{eqnarray}
}
Since the values of the variables are mutually constrained, it is not possible that, 
for instance, $u$ and $t$ have simultaneously their upper bound values (if $(1\menys K)/C \mp 1$
then $K/(1\menys C) \mg  1$ and viceversa). Any bound which is out of $[0,1]$ is not 
informative and the appropriate boundary, $0$ or $1$, is then used. Note that the
lower bound for $t$ will never be negative under the assumption $K\mg C$.
 
  Once we have established these bounds, we can use equation \ref{eq-x} to compute
the range for the real performance value of our tagger: $x$ will be minimum when 
 $u$ and $t$ are minimum, which produces the following bounds:
{\footnotesize
\begin{eqnarray}
\label{eq-xmin}
 \hspace*{-5mm}  x_{min} &\igual & K \menys C p \\
\label{eq-xmax}
 \hspace*{-5mm}  x_{max} &\igual & \left\{ \begin{array}{ll}
                     K \mes  C & \mbox{if $K\mpi 1\menys C$} \\
                     1\menys \frac{K+C-1}{p} & \mbox{if $K\mgi 1\menys C$}
                    \end{array}\right.
\end{eqnarray}
}
  As an example, let's suppose we evaluate a tagger on a test corpus
which is known to contain about $3\%$ of errors ($C\igual 0.03$), and obtain 
a reported performance of $93\%$\footnote{This is a realistic case
  obtained by~\cite{Marquez97a} tagger. Note that $93\%$ is the
  accuracy on ambiguous words (the equivalent overall accuracy was
  about 97\%).} ($K\igual 0.93$). In this case, equations 
\ref{eq-xmin} and \ref{eq-xmax} yield a range for the real performance $x$ 
that varies from $[0.93,0.96]$ when $p\igual 0$ to $[0.90,0.96]$ when $p\igual 1$.

   This results suggest that although we observe a performance of $K$, we
can not be sure of how well is our tagger performing without taking into
account the values of $t$, $u$ and $p$.

   It is also obvious that the intervals in the above example are too wide,
since they consider all the possible parameter values, even when they correspond
to very unlikely parameter combinations\footnote{For instance, it is not reasonable
that $u\igual 0$, which would mean that the tagger {\bf never} disambiguates
correctly a wrong word in the corpus, or $p\igual 1$, which would mean that
it {\bf always} makes the same error when both are wrong.}.
  In section \ref{seccio-restriccions} we will try to narrow those
intervals, limiting the possibilities to {\sl reasonable} cases.

\section{Reasonable Bounds for the Basic Parameters}
\label{seccio-restriccions}

  In real cases, not all parameter combinations will be equally likely.
In addition, the bounds for the values of $t$, $u$ and $p$ are closely
related to the similarities between the training and test corpora.
That is, if the training and test sets are extracted from the same 
corpus, they will probably contain the same kind of errors in the
same kind of situations. This may cause the training procedure to 
{\sl learn} the errors --especially if they are systematic-- and thus
the resulting tagger will tend to make the same errors that appear in 
the test set. On the contrary, if the training and test sets come
from different sources --sharing only the tag set-- the behaviour of the
resulting tagger will not depend on the right or wrong tagging of the
test set.

  We can try to establish narrower bounds for the parameters than those
obtained in section~\ref{seccio-plantejament}. 

  First of all, the value of $t$ is already constrained enough, due to
its high contribution ($1\menys C$) to the value of $K$, which forces 
$t$ to take a value close to $K$. For instance, applying the
boundaries in equation~\ref{eq-tbounds} to the case $C\igual 0.03$
and $K\igual 0.93$, we obtain that $t$ belongs to $[0.928,0.959]$. 

  The range for $u$ can be slightly narrowed considering the following: In the case
of independent test and training corpora, $u$ will tend to be equal to $t$. Otherwise,
the more biased towards the corpus errors is the language model, the lower $u$ will be. 
Note than $u\mg t$ would mean that the tagger disambiguates {\sl better} the 
noisy cases than the correct ones. 
Concerning to the lower bound, only in the case that all the errors in the training
and test corpus were systematic (and thus can be learned) could $u$ reach zero. However,
not only this is not a likely situation, but also requires a perfect--learning tagger. 
It seems more reasonable that, in normal cases, errors will be random, and the 
tagger will behave randomly on the noisy occurrences. This yields a lower bound for 
 $u$ of $1/a$, being $a$ the average ambiguity ratio for ambiguous words.

The {\sl reasonable} bounds for $u$ are thus
\vspace{-0.05cm}
$$
{\footnotesize
\frac{1}{a} \mpi u\mpi min\left\{t,\frac{1\menys K}{C}\right\}
}
\vspace{-0.05cm}
$$

  Finally, the value of $p$ has similar constraints to those of $u$. If the test 
and training corpora are independent, the probability of making the
same error, given that both are wrong, will be the random $1/(a\menys 1)$.  
If the corpora are not independent, the errors that
can be learned by the tagger will cause $p$ to rise up to (potentially) $1$. Again,
only in the case that all errors where systematic, could $p$ reach $1$.

Then, the {\sl reasonable} bounds for $p$ are:
\vspace{-0.05cm}
$$
{\footnotesize
max\left\{\frac{1}{a\menys 1},\frac{K\mes C\menys 1}{C}\right\} \mpi p\mpi 1
}
\vspace{-0.05cm}
$$

\section{On Comparing Tagger Performances}
\label{seccio-comparacio}
As stated above, knowing which are the {\sl reasonable} limits for the $u$,
$p$ and $t$ parameters enables us to compute the range in which
the real performance of the tagger can vary.

So, given two different taggers $T_1$ and $T_2$, and provided we know the
values for the test corpus error rate and the observed performance of
both cases ($C_1$, $C_2$, $K_1$, $K_2$), we can compare them by matching the
{\sl reasonable} intervals for the respective real performances $x_1$
and $x_2$. 

From a conservative position, we cannot strongly state than one of the
taggers performs better than the other when the two intervals overlap,
since this implies a chance that the real performances of both taggers
are the same.

 The following real example has been extracted from \cite{Marquez97a}:
The tagger $T_1$ uses only bigram information and has an observed 
performance on ambiguous words $K_1\igual 0.9135$ ($96.86\%$ overall). 
The tagger $T_2$ uses trigrams and automatically acquired context 
constraints and has an accuracy of $K_2 \igual 0.9282$ ($97.39\%$
overall). Both taggers have been evaluated on a corpus {\sc (wsj)} 
with an estimated error rate\footnote{The {\sc (wsj)} corpus
                                      error rate is estimated over all
                                      words. We are assuming that the
                                      errors distribute uniformly among
                                      all words, although  
                                      ambiguous words probably have a higher
                                      error rate. Nevertheless, a higher value for
                                      $C$ would cause the intervals to
                                      be wider and to overlap even
                                      more.} $C_1\igual C_2\igual 0.03$.
The average ambiguity ratio of the ambiguous words in the corpus 
is $a\igual 2.5$ tags/word.

These data yield the following range of {\sl reasonable} intervals for
the real performance of the taggers.
\begin{table}[htb] \centering
\begin{tabular}{c|c}
       for $p_i\igual (1/a)\igual 0.4$  &    for $p_i\igual 1$ \\ \hline
    $x_1 \in [91.35,94.05]$ & $x_1 \in [90.75,93.99]$\\
    $x_2 \in [92.82,95.60]$ & $x_2 \in [92.22,95.55]$
\end{tabular}
\end{table}

The same information is included in figure~\ref{f-intervals} which
presents the reasonable accuracy intervals for both taggers, for $p$
ranging from $1/a\igual 0.4$ to 1 (the shadowed part corresponds to the
overlapping region between intervals).

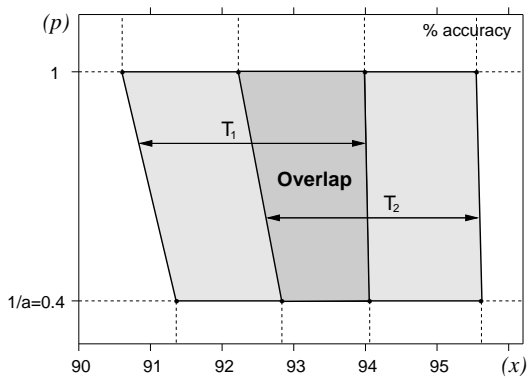
\figeps{intervals}{Reasonable intervals for both taggers}{f-intervals}

  The obtained intervals have a large overlap region which implies that
there are {\em reasonable} parameter combinations that could cause the
taggers to produce different observed performances though their
real accuracies were very similar. From this conservative approach, we
would not be able to conclude that the tagger $T_2$ is better than
 $T_1$, even though the $95\%$ confidence intervals for the observed 
performances did allow us to do so.

\section{Discussion}
\label{seccio-conclusions}

 The presented analysis of the effects of noise in the test corpus
on the evaluation of {\sc pos} taggers leads us to conclude that 
when a tagger is evaluated as better than another using noisy test
corpus, there are {\sl reasonable} chances that they are in fact 
very similar but one of them is just adapting better than the other
to the noise in the corpus.

 We believe that the widespread practice of evaluating taggers against
a noisy test corpus has reached its limit, since the performance
of current taggers is getting too close to the error rate usually
found in test corpora.

 An obvious solution --and maybe not as costly as one might think,
since small test sets properly used may yield enough statistical evidence--
is using only error--free test corpora. Another possibility is to 
further study the influence of noise in order to establish a
criterion --e.g. a threshold depending on the amount of
overlapping between intervals-- to decide whether a given tagger can be
considered better than another.

 There is still much to be done in this direction. This paper does
not intend to establish a new evaluation method for {\sc pos} tagging, but
to point out that there are some issues --such as the noise in test
corpus-- that have been paid little attention and are more important
than what they seem to be.

  Some of the issues that should be further considered are: The effect
of noise on unambiguous words; the reasonable intervals for {\sl overall} 
real performance; the --probably-- different values 
of $C$, $p$, $u$ and $t$ for ambiguous/unambiguous words; how to
estimate the parameter values of the evaluated tagger in order to
constrain as much as possible the intervals; the statistical
significance of the interval overlappings; a more informed (and 
less conservative) criterion to reject/accept the hypothesis that
both taggers are different, etc.

%


\newpage

\begin{other-abstract}[Resum]
  Aquest article versa sobre l'avaluaci\'o de desambiguadors
morfosint\`actics. Normalment, l'a\-va\-lu\-a\-ci\'o es fa comparant
la sortida del desambiguador amb un corpus de refer\`encia, que
se suposa lliure d'errors. De tota manera, els corpus que 
s'usen habitualment contenen soroll que causa que el rendiment
que s'obt\'e dels desambiguadors sigui una distorsi\'o del
valor real. En aquest article analitzem fins a quin punt a\-ques\-ta
distorsi\'o pot invalidar la comparaci\'o entre desambiguadors o
la mesura de la millora aportada per un nou sistema.
La conclusi\'o principal \'es que cal establir procediments
alternatius d'ex\-pe\-ri\-men\-ta\-ci\'o m\'es rigorosos, per poder a\-va\-lu\-ar
i comparar fiablement les precisions dels desambiguadors 
morfosint\`actics.
\end{other-abstract}

\begin{other-abstract}[Laburtena]
Artikulu hau desanbiguatzaile morfosintaktikoen ebaluazioaren inguruan
datza. Normalean, ebaluazioa, desanbiguatzailearen irte\-era eta ustez 
errorerik gabeko erreferentziako corpus bat konparatuz egiten da.
Hala ere, maiz corpusetan erroreak egoten dira eta ho\-rrek
desanbiguatzailearen emaitzaren benetako balioan eragina izaten du.
Artikulu honetan, hain zuzen ere, horixe aztertuko dugu, alegia, zer
neurritan distortsio horrek jar dezakeen au\-zitan desanbiguatzaileen arteko
konparazioa edo sistema berri batek ekar dezakeen hobekuntza-maila.
Konklusiorik nagusiena hauxe da: desanbiguatzaile morfosintaktikoak
aztertzeko eta modu ziurrago batez konparatu ahal izateko, azterketa-bideak
sakonagoak eta zehatzagoak izan beharko liratekeela.
\end{other-abstract}

\begin{thebibliography}{fullname}
{\small
{\parskip 0pt


\bibitem[\protect\citename{Church}1992]{Church92} Church,~K.W. 
\newblock 1992.
\newblock Current Practice in Part of Speech Tagging and Suggestions for the Future.
\newblock In Simmons (ed.), {\em Sbornik praci: In Honor of Henry Ku\v{c}era.} Michigan Slavic Studies.





\bibitem[\protect\citename{Krovetz}1997]{Krovetz97} Krovetz,~R.
\newblock 1997.
\newblock Homonymy and Polysemy in Information Retrieval.
\newblock In {\em Proceedings of joint E/ACL meeting.}

\bibitem[\protect\citename{M\`arquez and Padr\'o }1997]{Marquez97a}
        M\`arquez,~L. and Padr\'o,~L.
\newblock 1997.
\newblock A Flexible POS Tagger Using an Automatically Acquired Language Model.
\newblock In {\em Proceedings of joint E/ACL meeting.}

\bibitem[\protect\citename{Mooney}1996]{Mooney96} Mooney,~R.J.
\newblock 1996.
\newblock Comparative Experiments on Disambiguating Word Senses: An 
Illustration of the Role of Bias in Machine Learning.
\newblock In {\em Proceedings of EMNLP'96 conference.}

\bibitem[\protect\citename{Samuelsson and Voutilainen}1997]{Samuelsson97} 
        Samuelsson,~C. and Voutilainen,~A.
\newblock 1997.
\newblock Comparing a Linguistic and a Stochastic Tagger.
\newblock In {\em Proceedings of joint E/ACL meeting.}

}
}
\end{thebibliography}
\end{document}